\documentclass[11pt]{article}
\usepackage{geometry}
\usepackage{coling2020}
\usepackage{times}
\usepackage{url}
\usepackage{latexsym}
\usepackage{microtype}
\usepackage{graphicx}
\graphicspath{ {./images/} }
\colingfinalcopy 

\title{CS-NLP team at SemEval-2020 Task 4: Evaluation of State-of-the-art NLP Deep Learning Architectures on Commonsense Reasoning Task}

\author{Sirwe Saeedi \\
  Western Michigan University, USA \\
  {\tt sirwe.saeedi@wmich.edu} \\\And
  Aliakbar Panahi \\
  Virginia Commonwealth University, USA\\
  {\tt panahia@vcu.edu} \\ \AND
  Seyran Saeedi \\
  Virginia Commonwealth University, USA \\
  {\tt saeedis@vcu.edu} \\\And
  Alvis C Fong \\
  Western Michigan University, USA \\
  {\tt alvis.fong@wmich.edu} }
\date{}

\begin{document}
\maketitle
\begin{abstract}
In this paper, we investigate a commonsense inference task that unifies natural language understanding and commonsense reasoning. We describe our attempt at SemEval-2020 Task 4 competition: Commonsense Validation and Explanation (ComVE) challenge. We discuss several state-of-the-art deep learning architectures for this challenge. Our system uses prepared labeled textual datasets that were manually curated for three different natural language inference subtasks. The goal of the first subtask is to test whether a model can distinguish between natural language statements that make sense and those that do not make sense. We compare the performance of several language models and fine-tuned classifiers. Then, we propose a method inspired by question/answering tasks to treat a classification problem as a multiple choice question task to boost the performance of our experimental results (96.06\%), which is significantly better than the baseline. For the second subtask, which is to select the reason why a statement does not make sense, we stand within the first six teams (93.7\%) among 27 participants with very competitive results. Our result for last subtask of generating reason against the nonsense statement shows many potentials for future researches as we applied the most powerful generative model of language (GPT-2) with 6.1732 BLEU score among first four teams\footnote{\url{https://github.com/Sirwe-Saeedi/Commonsense-NLP}}.

  \end{abstract}
\section{Introduction\blfootnote{This work is licensed under a Creative Commons Attribution 4.0 International Licence. Licence details: http://creativecommons.org/licenses/by/4.0/}}

\label{section:1}
Commonsense is unstated background knowledge that is used to perceive, infer, and understand the physical world, human emotions, reactions, and knowledge of the common facts that most people agree with. Ordinary commonsense helps us to differentiate between simple false and true statements or answer questions, such as ``can an elephant fit into the fridge'' quickly, but they can be difficult for automatic systems \cite{c206796aef5c4a1e8fda075d6fd94673}. Recent advances in machine learning emphasize the importance of commonsense reasoning in natural language processing (NLP) and as a critical component of artificial intelligence (AI). In the fifty-year history of AI research,  progress was extremely slow \cite{4aba22e1f5b0492bab5811af4028ff48} in automated commonsense reasoning. However, when transfer learning ~\cite{Yosinski2014HowTA,Goodfellow-et-al-2016} and then transformers were introduced to the NLP world \cite{vaswani1706attention}, great breakthroughs and developments have occurred at an unprecedented pace \cite{pan2009survey,Tan2018ASO}. Advances in machine learning and deep learning methods have been achieved in numerous studies and wide range of disciplines \cite{Panahi2019word2ketSW,nemati2020machine,article,rs12091361,arodz2019quantum,oh2018effects}.

This paper describes a system participating in the SemEval-2020 ``Commonsense Validation and Explanation (ComVE) Challenge'', multiple tasks of commonsense reasoning and Natural Language Understanding (NLU) designed by \cite{wang-etal-2020-semeval}. The competition is divided into three subtasks, which involve testing commonsense reasoning in automatic systems, multiple choice questions, and text generation. In these tasks, participants are asked to improve the performance of previous efforts \cite{wang2019does}. We apply statistical language modeling, or language modeling (LM) for short as one of the most important parts of modern NLP and then transfer learning to reuse a pretrained model on different data distribution and feature space as the starting point of our target tasks. Applying Transfer Learning to NLP significantly improves the learning process in terms of time and computation through the transfer of knowledge from a related task that has already been learned \cite{10.5555/1803899}.

Language modeling is the task of probability distribution over sequences of words. It also assigns a probability for the likelihood of a given word (or a sequence of words) to follow a sequence of words. Language modeling are applied to many sorts of tasks, like: Machine Translation, Speech Recognition, Question Answering, Sentiment analysis, etc. AWD-LSTM (ASGD Weight-Dropped LSTM) \cite{merity2017regularizing} is a fundamental building block of language modeling which uses different gradient update step and it returns the average value of weights in previous iterations instead of current iteration.

Recently, there have been some excellent advancements towards transfer learning, and its success was illuminated by Bidirectional Encoder Representations from Transformers (BERT) \cite{devlin2018bert}, OpenAI transformer (GPT-2) \cite{radford2019language}, Universal Language Model Fine-tuning for Text Classification (ULMFiT) by fast.ai founder Jeremy Howard \cite{howard2018universal}, ELMo \cite{Peters:2018}, and other new waves of cutting-edge methods and architectures like XLNet \cite{yang2019xlnet}, Facebook AI RoBERTa: A Robustly Optimized BERT Pretraining Approach \cite{liu2019roberta}, ALBERT: A Lite BERT for Self-supervised Learning of Language Representations \cite{lan2019albert}, T5 team google \cite{raffel2019exploring}, and CTRL \cite{keskar2019ctrl}. For this work, we employ and fine-tune some of these suitable models.

When BERT was published, it achieved state-of-the-art performance on a number of natural language understanding tasks. As opposed to directional models like word2vec \cite{mikolov2015computing}, which generates a single word representation for each word in the vocabulary and read the text input sequentially, BERT is deeply bidirectional and reads the entire sequence of words at once. Therefore, BERT allows the model to learn the context of a word based on all of its surroundings using two training strategies: Masked Language Model (MLM) and Next Sentence Prediction (NSP). MLM technique is masking out some of the words in the input and then condition each word bidirectionally to predict the masked words. A random sample of the tokens in the input sequence is selected and replaced with the special token `[MASK]' and the objective is a cross-entropy loss on predicting the masked tokens \cite{devlin2018bert}. In this paper, we use a method inspired by MLM. 

The paper ``Attention Is All You Need'' \cite{vaswani1706attention} describes a sequence-to-sequence architecture called transformers relying entirely on the self-attention mechanism and does not rely on any recurrent network such as GRU \cite{chung2014empirical} and LSTM \cite{hochreiter1997long}. Transformers consist of Encoders and Decoders. The encoder takes the input sequence and then decides which other parts of the sequence are important by attributing different weights to them. Decoder turns the encoded sentence into another sequence corresponding to the target task.

A huge variety of downstream tasks have been devised to test a model's understanding of a specific aspect of language. The General Language Understanding Evaluation (GLUE) \cite{wang2018glue} and The Natural Language Decathlon (decaNLP) benchmarks \cite{mccann2018natural} consist of difficult and diverse natural language task datasets. These benchmarks span complex tasks, such as question answering, machine translation, textual entailment, natural language inference, and commonsense pronoun resolution. The majority of state-of-the-art transformers models publish their results for all tasks on the GLUE benchmark. For example,  models like different modified versions of BERT, RoBERTa, and T5 outperform the human baseline benchmark \cite{zhu2019freelb,wang2019structbert}. For the evaluation phase, GLUE follows the same approach as SemEval.

Our attempts at SemEval-2020 Task4 competition boost performance on two subtasks significantly. Our idea in reframing the first subtask helps to outperform results of state-of-the-art architecture and language models like BERT, AlBERT, and ULMFiT. General-Domain of ULMFiT is to predict the next word in a sequence by a widely used pretrained AWD-LSTM network on the WikiText-103 dataset. ULMFiT could outperform many text classification tasks like emotion detection, early detection of depression, and medical images analysis \cite{lundervold2019overview,xiao2019figure,trotzek2018utilizing}. We were ranked 11 with a very competitive result on the first subtask and achieved rank 6 for the second subtask amongst 40 and 27 teams, respectively. 
 
This paper is organized as follows. In Section 2, we introduce the three subtasks and their datasets. In Section 3, we describe our different applied models and various strategies that were used to fine tune the models for each individual subtask. In Section 4, we present the performance of our system. Finally, we conclude the paper in Section 5.

\section{Task Definition and Datasets}
\begin{figure}[ht]
\includegraphics[width=9cm, height=9cm]{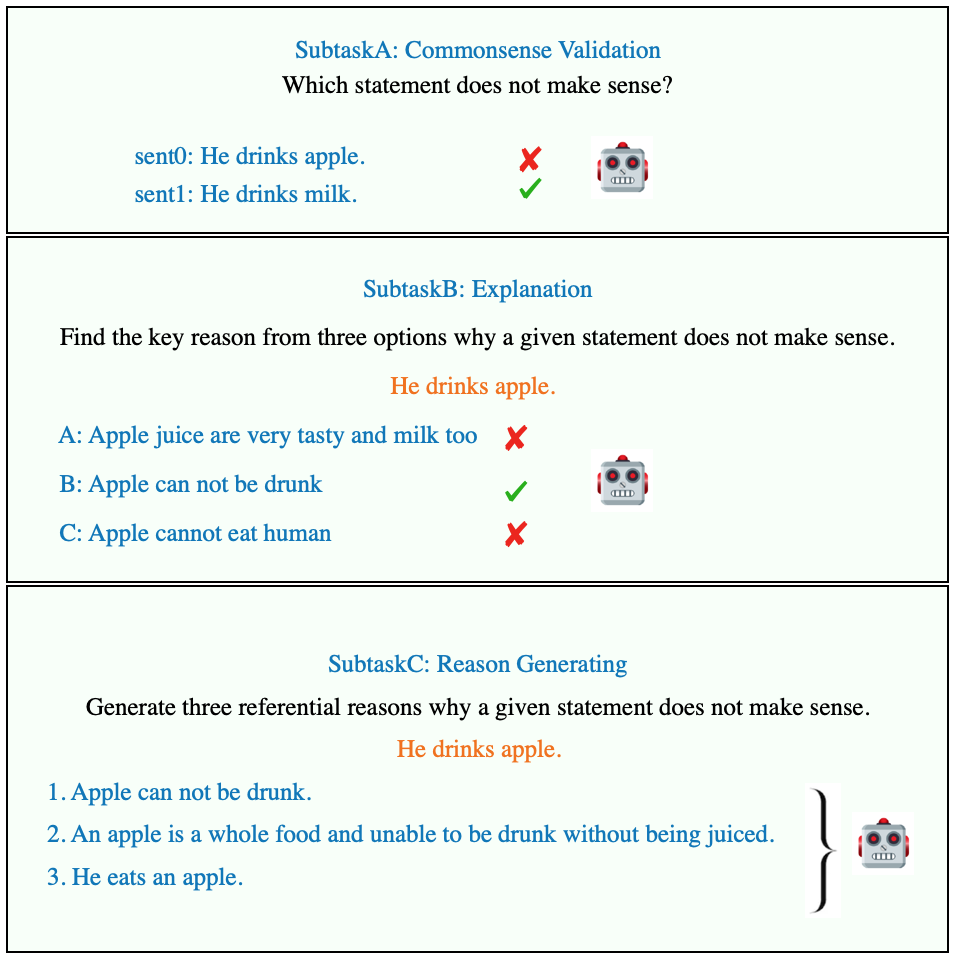}
\centering
\caption{Sample of training data for each subtask}
\label{fig:Figure1}
\end{figure}

As discussed, SemEval-2020 Task 4 consists of three subtasks, each designed for a different natural language inference task. Figure \ref{fig:Figure1} shows a sample for each subtask and the corresponding answer of the model.

\begin{itemize}
\item {\bf SubtaskA (Commonsense Validation)}: Given two English statements with similar wordings, decide  which one does not make sense. We had access to 10,000, and 2021 human-labeled pairs of sentences for training and trial models, respectively. After releasing the dev set, we combined these two datasets for training phase and used the dev set to test our models.
\item {\bf SubtaskB (Explanation)}: Given the nonsense statement, select the correct option among three to reason why the statement conflicts with human knowledge. The number of samples in the datasets for this task is similar to subtaskA. Each sample contains the incorrect statement from subtaskA and three candidate reasons to explain why this is against commonsense knowledge.
\item {\bf SubtaskC (Reason Generating)}: Given the nonsense statement, generate an understandable reason in the form of a sequence of words to verify why the statement is against human knowledge. Training samples of datasets for this subtask are all of the false sentences in subtaskA as well as for trial and dev set.
\end{itemize}

\section{Model Description}
Large pretrained language models are definitely the main trend of the latest NLP breakthroughs. As transformers occupy the NLP leaderboards, we choose several state-of-the-art architectures to outperform the baseline of all subtasks significantly. For each subtask, we describe our system separately below. 

\subsection{SubtaskA (Commonsense Validation)}

We consider two approaches to address this task: the first method is based on language models, and the second approach uses classifiers. Our experimental process begins with language models which the key idea behind the first approach is to find the probability of appearing each word in statements and then select one with higher multiplication of probabilities.

Our first try involves fine-tuning pretrained model on AWD-LSTM (as described in Section \ref{section:1}), which performs as poor as a random guess. We also try two other different language models: `BERT' the MLM that attempts to predict the original value of the masked words, based on the non-masked words in the sequence of words, and then transformer.

Original BERT uniformly selects 15\% of the input tokens for possible replacement. Of the selected tokens, 80\% are replaced with `[MASK]', 10\% are left unchanged, and 10\% are replaced by a randomly selected vocabulary token. However, our way of using MLM follows these steps: 
\begin{enumerate}
	\item Add  special tokens to the beginning and end of each sentence. \newline [`[CLS]', `He', `drinks', `apple', `[SEP]']
	\item Replace each token from left to right by `[MASK]' each time. 
	
	[`[MASK]', `He', `drinks', `apple', `[SEP]'],
	
	[`[CLS]', `[MASK]', `drinks', `apple', `[SEP]'], 
	
	[`[CLS]', `He', `[MASK]', `apple', `[SEP]'],  
	
	[`[CLS]', `He', `drinks', `[MASK]', `[SEP]'],
	
	[`[CLS]', `He', `drinks', `apple', `[MASK]']
	\item Feed them to MLM for predicting the probabilities of the original masked tokens.
	\item Normalize predicted probabilities using softmax activation function in the output layer.
	\item Multiply predicted probabilities of masked tokens for each pair of statement. The correct sentence has a higher probability.
\end{enumerate}
During the consideration of dataset homogeneity, we observe some samples are ended by periods and some others are not. The most frequent reason for using periods is to mark the end of sentences that are not questions or exclamations. By adding a period at the end of all statements, we increase the accuracy by 4\%, which is remarkable. We also try normalization and padding to boost performance of the model to minimize the impact of sequence length. Surprisingly, during normalization in step 4 we observe that normalizing by the length of sequence root of multiplied probabilities does not improve the performance of model. Similarly, normalization using perplexity to evaluate language models does not increase the accuracy of model. Perplexity is the inverse probability of the test set, normalized by the number of words and minimizing perplexity is the same as maximizing probability. We observe that padding does not make any differences in terms of accuracy. Therefore, the result of BERT MLM is almost the same with the baseline, which is achieved by fine-tuned ELMo as reported by \cite{wang2019does}. As a result, our observation shows BERT MLM model is more suitable for long document understanding; however, the maximum length (27) of our samples is too short.  

As mentioned, we consider classifiers as the second approach to deal with this task. We show that the classification-based approach is more efficient in recognizing nonsense statements except ULMFiT for text classification. Our main reasons for applying ULMFiT to address subtaskA are its techniques to deal with a small domain of dataset: Discriminative fine-tuning, slanted triangular learning rates instead of using the same learning rate throughout training, and gradual unfreezing neural network layers. However, applying ULMFiT for this task is similar to choosing between any two statements, randomly. On the other hand, our results show the fine-tuned classifier on the pretrained AWD-LSTM, transformer, and random guess yielded results with almost close to 50\% accuracy. As shown in Table 1, these models can not differentiate sentences that make sense from those that do not make sense, properly.

In addition, we apply the ubiquitous architecture of transformers for classification, such as BERT, Albert, and RoBERTa. All these models allow us to pretrain a model on a large corpus of data, such as all Wikipedia articles and English book corpus, and then fine-tune them on downstream tasks. Looking at Table 1, we see RoBERTa outperforms all other models. We find out a significant difference when using fine-tuned Albert and BERT classification. Table \ref{table:kysymys} summarizes the performance of these systems on dev in terms of accuracy.

\begin{table}[ht]
\begin{center}
\begin{tabular}{ | l | l | l | p{5cm} |}
\hline \bf Models & \bf Accuracy \\ \hline
AWD-LSTM & 52.45 \\ \hline
Transformer & 53.8 \\ \hline
ULMFiT & 59.8     \\ \hline
BERT MLM & 74.29  \\ \hline
BERT classification & 88 \\ \hline
Albert classification & 92 \\ \hline
RoBERTa classification & 95 \\ \hline 
RoBERTa multiple choice question & 96.08\\ \hline
\end{tabular}
\end{center}
\caption{Experimental results for subtaskA on dev set. }
\label{table:kysymys}
\end{table}

Our idea to boost the performance of all these applied models is reframing the input of subtaskA as a binary classification task to the input of another downstream task, multiple choice questions. As a result, we show fine-tuned RoBERTa for multiple choice questions task gives better results than RoBERTa for classification problem on both dev and test set (See Table \ref{table:kysymys}). 

The difference between these two models is paying attention to the statements. In the self-attention layer, the encoder looks at other words in the input sentence as it encodes a specific word. For binary classification models like BERT, RoBERTa, and Albert, we concatenate two statements and then self-attention layer attends to each position in the input sequence, including both statements. However, for RoBERTa multiple choice questions task, we feed each statement to the network separately. Therefore, the attention layer attends to the sequence of words for each individual statement for gathering information that can lead to better encoding for each word. 

Question answering task usually provides a paragraph of context and a question. The goal is to answer the question based on the information in the context. For subtaskA, we do not have the context and question; all we have is two options corresponding to the statements which are fed to the network, separately. Our goal is to select the correct statement (answer) from the two options.

As expected, determining optimal hyper-parameters has a significant impact on the accuracy on the performance of the model, and their optimization needs careful evaluation of many key hyper-parameters. We primarily follow the default hyper-parameters of RoBERTa, except for the maximum sequence length, weight decay, and learning rate $\in\{1e-5, 2e-5, 3e-5\}$ which is warmed up over 320 steps with a maximum of 5336 numbers of step to a peak value and then linearly decayed. The other hyper-parameters remained as defaults during the training process for 5 epochs. By searching the hyperparameter space for the optimum values, fine-tuned hyper-parameters achieve 96.08\% and 94.7\% accuracy on dev and test set, respectively. Our result is a big jump from 74.1\% baseline accuracy and competes with 99.1\% accuracy of human performance.

\subsection{SubtaskB (Explanation)}
As described earlier, subtaskB requires world knowledge and targets commonsense reasoning to answer why nonsense statements do not make sense. This type of task seems trivial for humans with a basic knowledge but is still one of the most challenging tasks in the NLP world. However, the baseline for human performance, 97.8\% shows how it is difficult to reason even with a comprehensive commonsense knowledge. 

Our goal is to investigate whether transformers like RoBERTa (which its performance was confirmed on subtaskA) can learn commonsense inference given a nonsense statement. The architecture of RoBERTa-large is comprised of 24-layer, 1024-hidden dimension, 16-self attention heads, 355M parameters and pretrained on book corpus plus English Wikipedia, English CommonCrawl News, and WebText corpus.  

SubtaskB is a multiple choice question task and we fine-tune hyper-parameters of RoBERTa model to answer questions. In this setting, we concatenate the nonsense statement (context) with each option (endings) and then use three statements as the input of model. For example, `He drinks apple.' is the context and [`Apple juice are very tasty and milk too.', `Apple can not be drunk.', `Apple cannot eat a human.'] is the list of endings. We want to select the ending from three options that is entailed by the context: 
\begin{itemize}
\item ``He drinks apple. Apple juice are very tasty and milk too.''
\item ``He drinks apple. Apple can not be drunk.''
\item ``He drinks apple. Apple cannot eat a human.''
\end{itemize}
The set of concatenated examples is fed into the model to predict the answer of questions that require reasoning. We considered a few hyper-parameter settings and figured out the model with hyper-parameters in Table \ref{Table 2} yields the surprising results 93.7\%, compared to the baseline accuracy of 45.6\%.

\begin{table}[ht]
\begin{center}
\begin{tabular}{ | l | l | l | p{8cm} |}
\hline \bf hyper-parameters & \bf value \\ \hline
batch size & 16  \\ \hline
learning rate & $1e-5$ \\ \hline
weight decay & 0.1 \\ \hline
adam epsilon & $1e-8$   \\ \hline
num\_train\_epochs & 5  \\ \hline
max\_steps & 5336 \\ \hline
warmup\_steps & 320 \\ \hline
\end{tabular}
\end{center}
\caption{Tuned hyper-parameters of RoBERTa for subtaskB.}
\label{Table 2}
\end{table}

\subsection{SubtaskC (Reason Generating)}
Based on the subtaskC definition, we can frame subtaskC as a conditional text generation problem. Given a nonsense statement, we expect that the language model will generate commonsense reasons to explain why statement conflicts with our knowledge. We applied the full version of OpenAI GPT2 (Generative Pre-Training), a large-scale unsupervised language model with billions of parameters, trained on a very large corpus of text data. The goal of this model is to automatically generate text, given a sequence of natural language words. The performance of GPT-2 in a zero-shot setting is competitive on many language modeling datasets and various tasks like reading comprehension, translation, and question answering.

GPT-2 architecture claims that the model performs well in generating coherent samples depending on the context, which are fairly represented during the training process. However, we observed that employing GPT-2 for generating texts against the given nonsense statements is poor in performance with unnatural topic switching and 6.1732 BLEU score. We used the Pytorch implementation of GPT-2 (with all default hyper-parameters) that is provided by Huggingface transformers \cite{wolf2019huggingface} for natural language generation.

The GPT-2 is built using transformer decoder blocks. The key behined GPT-2 is called “auto-regression” that outputs one token at a time and after each token is produced, that token is added to the sequence of inputs then the new produced sequence becomes the input to the model.

Notably, we submitted the original test set (including nonesense statements) for the evaluation phase on SemEval-2020 portal and surprisingly, we stand among the first four teams. The competitive BLEU score of 17.2 with the top team shows that subtaskC is challenging enough to receive more research attentions. We believe that our simple and naive efforts indicate significant opportunities for future research to utilize reasoning on commonsense knowledge.  

\section{Conclusion}
We evaluated architectures for three commonsense reasoning tasks. First, we found that RoBERTa-large performs better substantially in differentiating sentences that make sense from those that do not make sense compared to other cutting-edge architectures (e.g. Albert, BERT, and ULMFiT). We reframe this classification task to a question answering task to enhance the performance of the fine-tuned RoBERTa to 96.08\%. Second, we achieved significant results on reasoning why false statements do not make sense. We showed that RoBERTa performs well in selecting the correct option among three to infer the commonsense reason and it yields significant result with 93.7\% accuracy compare to baseline using BERT, 45.6\%. With a little effort on generating reasons to explain why false statement conflicts with commonsense knowledge, we observe that the original test set produces 17.2 BLEU score which ranked us among first four teams in the competition with a very competitive results. Our experimental result showed that GPT-2 performs as poor as random generating of a sequence of words for this task. We believe this task has many potentials and challenges for upcoming NLP researches. As another future work, we believe that ensemble learning can reduce the variance of predictions and also improve prediction performance. In ensemble learning, multiple models are generated and combined to address the subtasks and reduce the likelihood of an unfortunate selection of a poor one.

\bibliographystyle{coling}
\bibliography{semeval2020}
\end{document}